\documentclass{article}

 \PassOptionsToPackage{numbers, compress}{natbib}
\usepackage[final]{nips_2017}

\usepackage{hyperref}       
\usepackage{url}            
\usepackage{booktabs}       
\usepackage{amsfonts}       
\usepackage{nicefrac}       
\usepackage{microtype}      
\usepackage{enumitem}
\usepackage{tcolorbox}
\usepackage{cite}
\usepackage{float}		
\usepackage{amsfonts,epsfig,amsmath,latexsym,amssymb,amscd,multirow,graphicx,geometry,lscape,amsmath,amssymb,bm,pifont,graphicx,amssymb,amsmath}
\usepackage[autostyle]{csquotes} 
\usepackage{cancel,algorithm,algorithmic,algorithm,verbatim,array,multicol,palatino,amsfonts}

\usepackage{tabu,makecell}
\usepackage{graphicx}
\usepackage{lineno}
\usepackage{caption}
\usepackage{subcaption}
\usepackage[normalem]{ulem}
\usepackage{graphicx}
\usepackage{bbm}
\usepackage{dsfont}
\usepackage{amsthm}
\usepackage{amsmath}
\usepackage{arydshln}
\usepackage{mathrsfs}
\usepackage[framemethod=TikZ]{mdframed}
\usepackage{lipsum}
\usepackage{framed}

\newcommand{\x}{\mathbf{x}}

\newcommand{\Y}{\mathbf{Y}}
\newcommand{\B}{\mathbf{B}}

\newcommand{\V}{\mathbf{V}}

\newcommand{\exE}{\mathbb{E}}

\newcommand{\C}{\mathcal{C}}

\newcommand{\PSI}{\bm{\Psi}}
\newcommand{\THETA}{\bm{\theta}}

\newcommand{\xN}{\x_{\text{\tiny{1:N} } } }

\newcommand{\YN}{\Y_{\mathcal{N}  } }

\newcommand{\NA}{N_\text{\tiny{H}}}
\newcommand{\ND}{N_\text{\tiny{L}}}

\newcommand{\sigW}{\sigma^2_\text{\tiny{W}}}

\newcommand{\sigV}{\sigma^2_\text{\tiny{V}}}

\newcommand{\GP}{\mathcal{GP}}

\DeclareMathOperator*{\argmin}{arg\,min}

\renewcommand{\d}{\text{d}}

\newfont{\fsc}{eusm10}                         

\title{Sensor Selection and Random Field Reconstruction for Robust and Cost-effective Heterogeneous Weather Sensor Networks for the Developing World}

\author{
  Pengfei Zhang\\
  University of Oxford\\
  Oxford, UK \\
   \And
  Ido Nevat \\
  TUM CREATE \\
  Singapore \\
   \AND
  Gareth W. Peters \\
  Heriot-Watt University \\
  Scotland, UK \\
   \And
  Wolfgang Fruehwirt \\
  University of Oxford \\
  Oxford, UK \\
   \AND
  Yongchao Huang \\
  University of Oxford \\
  Oxford, UK \\
   \And
  Ivonne Anders \\
  ZAMG \\
  Vienna, Austria \\  
   \And
  Michael Osborne \\
  University of Oxford \\
  Oxford, UK \\
}

\begin{document}

\maketitle

\begin{abstract}
We address the two fundamental problems of \textit{spatial field reconstruction} and \textit{sensor selection} in heterogeneous sensor networks:
(i) how to efficiently perform \textit{spatial field reconstruction} based on measurements obtained simultaneously from networks with both high and low quality sensors; and (ii) how to perform \textit{query based sensor set selection with predictive MSE performance guarantee}.
For the first problem, we developed a low complexity algorithm based on the \textit{spatial best linear unbiased estimator} (S-BLUE).
Next, building on the S-BLUE, we address the second problem, and develop an efficient algorithm for \textit{query based sensor set selection with performance guarantee}. Our algorithm is based on the Cross Entropy method which solves the combinatorial optimization problem in an efficient manner.
\end{abstract}

\section{Introduction}

We consider the case where two types of sensors are deployed: the first consists of expensive, high quality sensors; and the second, of cheap low quality sensors, which are activated only if the intensity of the spatial field exceeds a pre-defined activation threshold (eg. wind sensors). This type of \textit{heterogeneous sensor networks} approach has gained attention in the last few years due to the vision of the Internet of Things (IoT) where networks may share their data over the internet \citep{gubbi2013internet,vermesan2011internet}.

Two practical scenarios \footnote{In particular, developing world countries are constrained by their budget when purchasing equipment. Meanwhile, these countries are heavily effected by climate change. This combination makes the robust and cost-effective sensor selection in weather sensor networks a major concern for the developing world.} that are of importance are:
firstly, high-quality sensors may be deployed by government agencies (eg. weather stations). These are sparsely deployed due to their high costs, limited space constraints, high power consumption etc. To improve the coverage of the WSN, low-quality cheap sensors can be deployed to augment the high-quality sensor network \citep{rajasegarar2014high}; Secondly, High-quality sensors cannot be easily deployed in remote locations, for example in oceans, lakes, mountains and volcanoes. In these cases, battery operated low-quality cheap sensors can be deployed \citep{werner2006deploying}.

More specifically, the following two fundamental problems are the focus of this paper: Firstly, \textit{\textbf{Spatial field reconstruction}:} the task is to accurately estimate and predict the intensity of a spatial random field, not only at the locations of the sensors, but at all locations \citep{peters2015utilize, nevat2015estimation, nevat2013random}, given heterogeneous observations from both sensor networks; Secondly, \textit{\textbf{Query based sensor set selection with performance guarantee}}: the task is to perform on-line sensor set selection which meets the QoS criterion imposed by the user, as well as minimises the costs of activating the sensors of these networks \citep{calvo2016sensor,joshi2009sensor,chepuri2015sparsity}.

\section{System model}
\label{system_model}

We now present the system model for the physical phenomenon observed by two types of networks.
\begin{enumerate}[noitemsep]
\item[A1] Consider a random spatial phenomenon (eg. wind) to be monitored defined over a $2$-dimensional space $\mathcal{X} \in \mathbb{R}^{2}$. The mean response of the physical process is a smooth continuous spatial function $f\left(\cdot\right):\mathcal{X} \mapsto  \mathbb{R}$, and is modelled as a Gaussian Process (GP) according to
\begin{align}
f\left(\x\right) \sim \GP\left(\mu_f \left(\x;\THETA_f\right)
,\C_f\left(\x_1,\x_2;\PSI_f\right)
\right),
\end{align}	
where the mean and covariance functions $\mu_f \left(\x;\THETA_f\right),\C_f\left(\x_1,\x_2;\PSI_f\right)$ are assumed to be known.
\item[A2] Let $N$ be the total number of sensors that are deployed over a $2$-D region $\mathcal{X} \subseteq \mathbb{R}^2$, with
$\x_{n} \in \mathcal{X}, n=\left\{1,\cdots, N\right\}$ being the physical location of the $n$-th sensor, assumed known by the FC. The number of sensors deployed by Network $1$ and Network $2$ are $\NA$ and $\ND$, respectively, so that $N=\NA+\ND$ .

\item[A3] \textbf{Sensor network $1$ includes high quality sensors.}
The sensors have a $0$-threshold activation and each of the sensors collects a noisy observation of the spatial phenomenon $f\left(\cdot\right)$. At the $n$-th sensor, located at $\x_n$, the observation is given by:
\begin{align}
Y^{H}\left(\x_n\right)= f\left(\x_n\right) + W\left(\x_n\right),  \;n=\left\{1,\cdots,\NA\right\}
\end{align}
where $W\left(\x_n\right)$ is i.i.d Gaussian noise $W\left(\x_n\right) \sim N\left(0,\sigW\right)$.
 \textbf{Sensor network $2$ includes low quality sensors.}
The sensors have a $T$-threshold activation and each of the sensors collects a noisy observation of the spatial phenomenon $f\left(\cdot\right)$, only if the intensity of the field at that location exceeds the pre-defined threshold $T$, (eg. anemometer sensors for wind monitoring \citep{adafruit,Anemo4403}). At the $n$-th sensor, located at $\x_n$, the observation is given by:
\begin{align}
Y^{L}\left(\x_n\right)=
\left\{
\begin{array}{ll}
f\left(\x_n\right)+V\left(\x_n\right),&f\left(\x_n\right)\geq T\\
V\left(\x_n\right),&f\left(\x_n\right)<T
\end{array}
\right.
\end{align}
where $V\left(\x_n\right)$ is i.i.d Gaussian noise $V\left(\x_n\right) \sim N\left(0,\sigV\right)$.

\end{enumerate}
%
\section{Field Reconstruction via Spatial Best Linear Unbiased Estimator (S-BLUE)}
\label{S-BLUE}
To perform inference in our Bayesian framework, one would typically be interested in computing the predictive posterior density at any location in space, $\x_* \in  \mathcal{X} $ , denoted $p\left(f_*|\YN\right)$. Based on this quantity, a point estimator, like the Minimum Mean Squared Error (MMSE) estimator can be derived:
\begin{align*}
\widehat{f_*}^{\text{MMSE}} = \int \limits_{-\infty}^{\infty}
p\left(f_*|\YN,\xN,\x_*\right) f_* \d f_*
\end{align*}

We develop the spatial field reconstruction via Best Linear Unbiased Estimator (S-BLUE), which enjoys a low computational complexity \citep{kay:1998}. The S-BLUE does not require calculating the predictive posterior density, but only the first two cross moments of the model.
The S-BLUE is the optimal (in terms of minimizing Mean Squared Error (MSE)) of all linear estimators and is given by the solution to the following optimization problem:
\begin{equation}
\label{S_BLUE_objective}
\widehat{f}_* :=\widehat{a} + \widehat{\B} \YN = \arg \min_{a, \B} \exE\left[\left(f_*- \left(a + \B \YN\right)\right)^2\right],
\end{equation}
where $\widehat{a} \in \mathbb{R}$ and $\widehat{\B} \in \mathbb{R}^{1 \times N}$.

The optimal linear estimator that solves (\ref{S_BLUE_objective}) is given by
\begin{align}
\begin{split}
\label{s_blue_estimate}
\hat{f_*}&=\exE_{f_*\; \YN}\left[f_* \; \YN\right]\exE_{\YN}\left[ \YN\; \YN\right]^{-1}\left(\YN-\exE\left[\YN\right]\right),
\end{split}
\end{align}
and the Mean Squared Error (MSE) is given by
\begin{align}
\begin{split}
\label{s_blue_estimate_MSE}
\sigma^2_{*}&=k\left(\x_*,\x_*\right)-
\exE_{f_*\; \YN}\left[f_*\; \YN\right]\exE_{\YN}\left[\YN\; \YN \right]^{-1}
\exE_{\YN\;f_*}\left[\YN \; f_*\right].
\end{split}
\end{align}

\section{Query Based Sensor Set Selection with Performance Guarantee}
\label{cross_entropy}
In this Section we develop an algorithm to perform on-line sensor set selection in order to meet the requirements of a query made by users of the system. In this scenario users can prompt the system and request the system to provide an estimated value of the spatial random field at a location of interest $\x_*$. 
We defined the activation sets of the sensors in both networks by
$\mathcal{S}_1 \in \left\{0,1\right\}^{\left|\NA \right|}, \mathcal{S}_2 \in \left\{0,1\right\}^{\left|\ND \right|}$. Then the sensor selection problem can be formulated as follows:
\begin{align}
\label{SSO}
\begin{split}
\mathcal{S} &= \argmin_{
\left(
\stackrel{\mathcal{S}_1 \in \left\{0,1\right\}^{\left|\NA \right|}}{\mathcal{S}_2 \in \left\{0,1\right\}^{\left|\ND \right|}}
\right)}
w_h \left|\mathcal{S}_1\right|+w_l \left|\mathcal{S}_2\right|,
\\
&\text{s.t.} \;\;\sigma^2_*< \sigma^2_q,
\end{split}
\end{align}
where $\sigma^2_q$ is the maximal allowed uncertainty at the query location $\x_*$, and $w_h$ and $w_l$ are the known costs of activating a sensor from Network $1$ and Network $2$, respectively.

Suppose we wish to maximize a function $U\left(\x\right)$ over some set $\mathscr{X}$. Let us denote the maximum by $\gamma^*$; thus,
\begin{align}
\label{CEMO}
\gamma^*=\max_{\x\in\mathscr{X}}U(\x).
\end{align}
The Cross Entropy Method (CEM) solves this optimization problem by casting the original problem (\ref{CEMO}) into an estimation problem of rare-event probabilities. By doing so, the CEM aims to locate an optimal parametric sampling distribution, that is, a probability distribution on $\mathscr{X}$, rather than locating the optimal solution directly. 
To apply the CEM to solve our optimization problem in (\ref{SSO}), we need to choose a parametric distribution. Since the activation of the sensors is a binary variable (eg. $0 \rightarrow \text{don't activate},1 \rightarrow \text{activate}$), we choose an independent Bernoulli variable as our parametric distribution, with a single parameter $p$ (ie. $\V=p$). The Bernoulli distribution is a member of the NEF of distributions, hence, an analytical solution of the stochastic program is available in closed form as follows:
\begin{align*}
p_{t,j} =\frac{
						\sum\limits_{i=1}^{K}
						\mathds{1}\left(\mathbf{\Gamma^H_{i,j}}=1\right)
						\mathds{1}\left(U\left(k\right) \geq \beta_t \right)
									}
						{\sum\limits_{i=1}^{K}
						\mathds{1}\left(U\left(k\right) \geq \beta_t \right)}.
\end{align*}
Since the optimization problem in Eq. (\ref{SSO}) is a constrained optimization problem, we introduce an Accept$\setminus$Reject step which rejects samples which do not meet the QoS criterion $\sigma^2_*< \sigma^2_q$, as follows
\begin{align*}
U\left(k\right)=
\begin{cases}
-\left(w_h \left|\mathcal{S}^H\right|+w_l \left|\mathcal{S}^L\right|\right),&\sigma^2_*\left(k\right) <\epsilon\\
-\infty,&\text{Otherwise}
\end{cases}
\end{align*}

\section{Experimental Results and Discussion}

\begin{figure}
\centering
\begin{subfigure}{.55\textwidth}
  \centering
  \includegraphics[height=4.5cm]{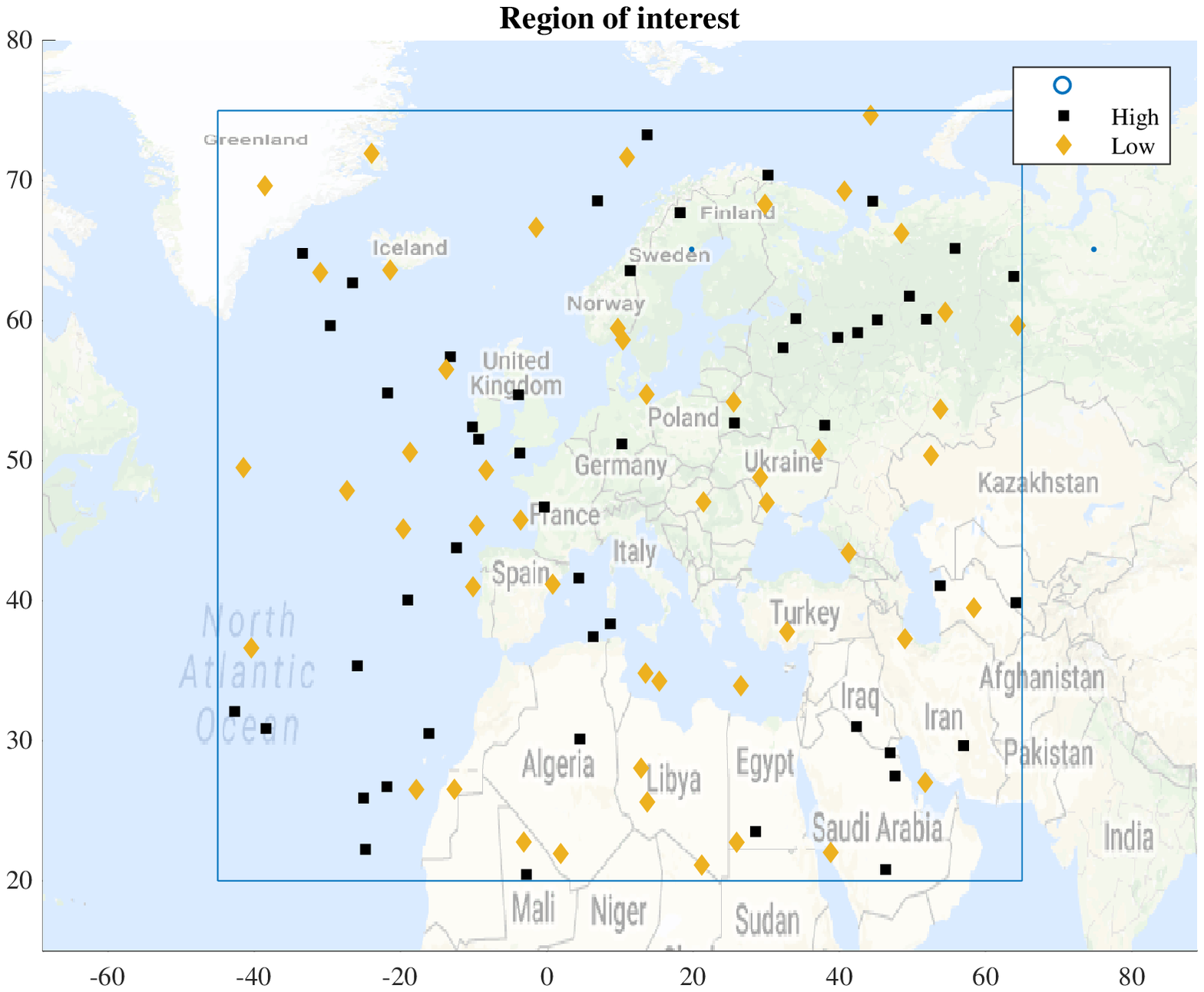}
\end{subfigure}%
\begin{subfigure}{.55\textwidth}
  \centering
  \includegraphics[height=4.5cm]{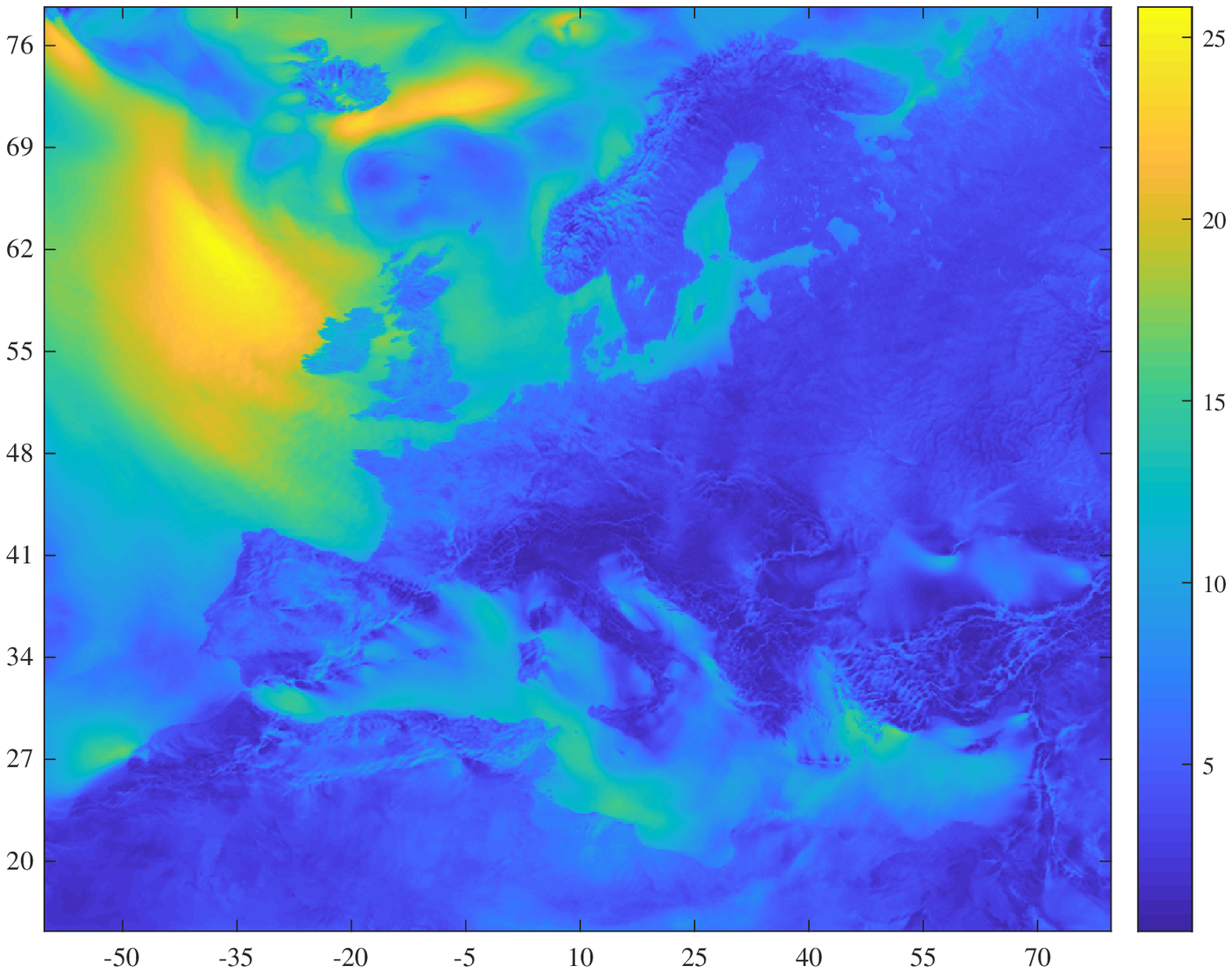}
\end{subfigure}
\caption{Left panel: map of region of interest with sensors locations. Right panel: Storm wind intensity map}			
\label{fig:google_true_storm}
\end{figure}

\begin{figure}
\centering
\begin{subfigure}{.55\textwidth}
  \centering
\includegraphics[height=4.5cm]{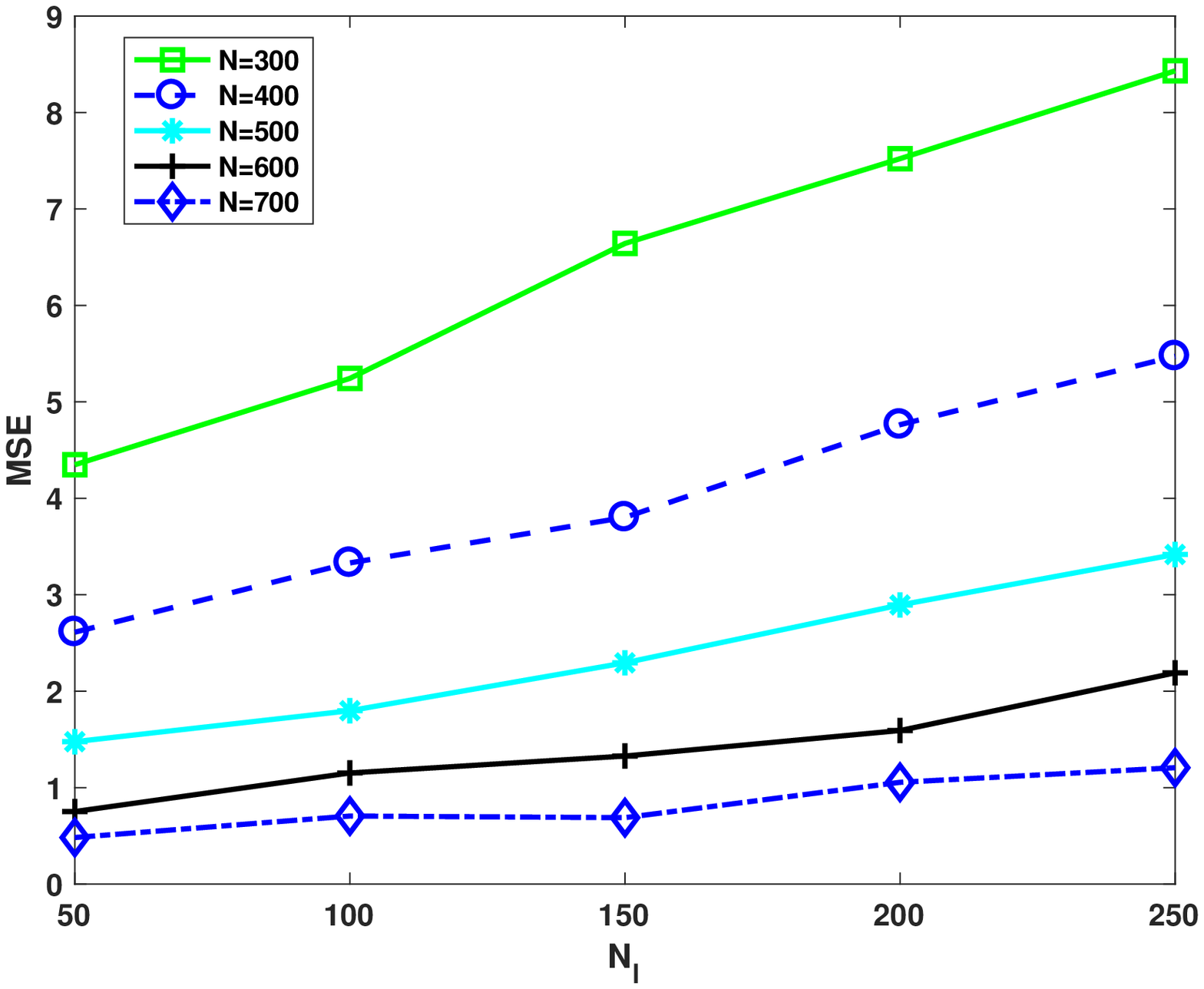}
\end{subfigure}%
\begin{subfigure}{.55\textwidth}
  \centering
\includegraphics[height=4.5cm]{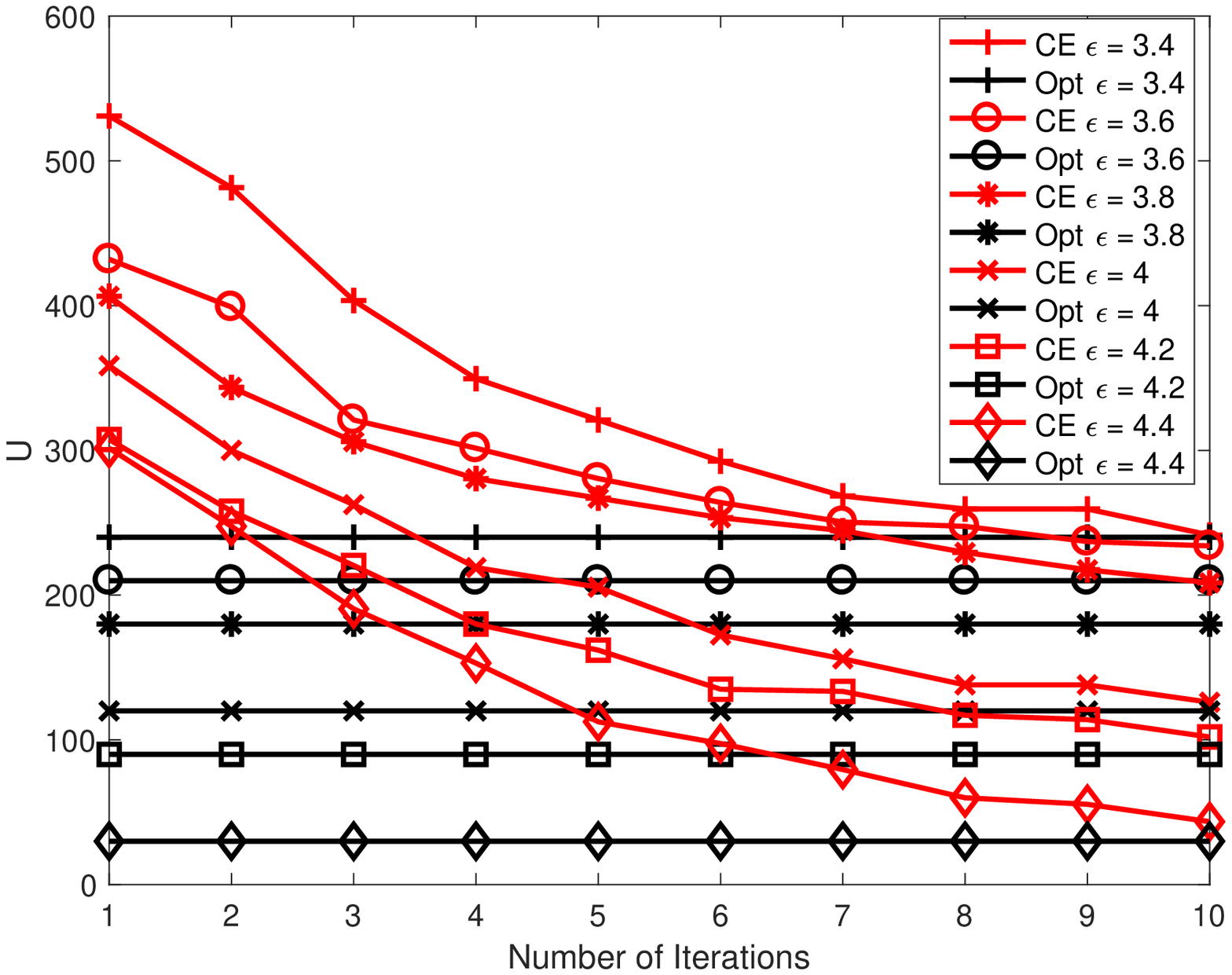}
\end{subfigure}
\caption{Left panel: MSE with effect of different number of high and low quality sensors. Right panel: Comparison of U values between optimal scheme and CE method with effect of number of iterations.}			
\label{fig:CompareRMSECost}
\end{figure}

In order to test our algorithm on real data sets, we use fine grained datasets available from Hans-Ertel-Centre for Weather Research (HErZ) project\footnote{https://www.herz-tb4.uni-bonn.de/index.php/hans-ertel-centre-for-weather-research/funding}. Particularly, we choose Feb 23, 2015 as testing data since it has the highest wind speed across the whole year. The left panel of Fig. \ref{fig:google_true_storm} shows the region of interest on the map. Both high and low quality sensors are selected randomly within the region. In this figure we randomly deployed $50$ high quality and $50$ low quality sensors. The right panel of Fig. \ref{fig:google_true_storm} shows daily wind speed intensity across the region. In left panel of Fig. \ref{fig:CompareRMSECost} we present a quantitative comparison of the MSE for various values of high and low quality sensors. The result shows a clear trend of MSE with the increasing of high and low quality sensors. We also illustrate how our sensor selection algorithm performs. For comparison, we use an optimal selection method which only selects the sensor set collections that minimize the U values and ensures that the QoS criterion is being met.
 The simulation parameters we have are: $\{N_h=5, N_l=10, T=0, w_h=150, w_l = 30, \sigma_w=0.001, \sigma_g=0.003, k_f(x_*,x_*)=5.8, x_*=10, y_*=50, \epsilon=\{3.4, 3.6, 3.8, 4, 4.2, 4.4\}\}$. We fix the $N_h=5, N_l=10$. The comparison is shown in right panel of Fig. \ref{fig:CompareRMSECost}. We also increase the number of iterations in CE method from 1 to 10. It shows CE method converges quickly to the optimal selection algorithm within 10 iterations for all the $\epsilon$ values.


% ...


\medskip

\small

\bibliographystyle{unsrt}
\bibliography{references}




\end{document}